\documentclass[letterpaper, 10 pt, conference]{ieeeconf}  

\IEEEoverridecommandlockouts                              

\usepackage{amsmath} 
\usepackage{amssymb}  
\usepackage{algorithm}
\usepackage[noend]{algorithmic}
\usepackage{booktabs}
\usepackage{cite}
\usepackage{graphicx}
\usepackage[caption=false, font=footnotesize]{subfig}
\usepackage{multirow}
\usepackage{xcolor}

\usepackage{xspace}

\renewcommand{\eqref}[1]{(\ref{#1})}

\newcommand{\ie}{\textit{\textrm{i.e.}}}
\newcommand{\eg}{\textit{\textrm{e.g.}}}

\def\cpnet{$\phi_{node}$}
\def\emnet{$\phi_{edge}$}
\def\epfunc{f_{ep}}
\def\gls{SLS}
\def\glnh{GLN-Heu}
\def\glne{SLS}
\def\glnea{SLS-Aug}
\def\glnhtab{\textcolor{orange}{GLN-Heu}}
\def\glnetab{\textcolor{cyan}{SLS}}
\def\glneatab{\textcolor{teal}{SLS-Aug}}
\def\easy{\textit{Easy}}
\def\med{\textit{Medium}}
\def\hard{\textit{Hard}}
\def\srloc{SR-HL}
\def\pcloc{PC-HL}
\def\srnav{SR-Nav}
\def\pcnav{PC-Nav}
\def\sam{SAM}
\def\samlong{Scene Action Map}
\def\hm{Hand}
\def\fp{Flr}
\def\sm{SatMap}
\def\hmgt{Hand-HA}
\def\fpgt{Flr-HA}

\def\weblink{\urllink[pre = \bgroup\bf, post = \egroup]}

\makeatletter
\let\NAT@parse\undefined
\makeatother
\usepackage{hyperref}

\title{\LARGE \bf
Scene Action Maps: Behavioural Maps for \\Navigation without Metric Information
}

\author{Joel Loo$^{1}$ and David Hsu$^{1}$
\thanks{$^{1}$School of Computing \& Smart Systems Institute, National University of
Singapore. {\tt\small \{joell, dyhsu\}@comp.nus.edu.sg}
}%
}

\begin{document}

\maketitle
\thispagestyle{empty}
\pagestyle{empty}

\begin{abstract}

Humans are remarkable in their ability to navigate without metric information. We can read abstract 2D maps, such as floor-plans or hand-drawn sketches, and use them to navigate in unseen rich 3D environments, without requiring prior traversals to map out these scenes in detail. We posit that this is enabled by the ability to represent the environment abstractly as interconnected \textit{navigational behaviours}, \eg{}, ``follow the corridor'' or ``turn right'',  while avoiding detailed, accurate spatial information at the metric level.  We introduce the \emph{\samlong{}} (\sam{}), a behavioural topological graph, and propose a learnable \textit{map-reading} method, which parses a variety of 2D maps into \sam{}s.  Map-reading extracts salient information about navigational behaviours from the  overlooked wealth of pre-existing, abstract and inaccurate maps, ranging from floor-plans to sketches. We evaluate the performance of \sam{}s for navigation, by building and deploying a behavioural navigation stack on a quadrupedal robot. Videos and more information is available at: \href{https://scene-action-maps.github.io}{https://scene-action-maps.github.io}.

\end{abstract}

\section{Introduction}
\label{sec:intro}

Can robots navigate with limited metric and spatial information, just as humans do? Currently, most robots' navigation systems rely on detailed geometric maps and accurate metric positioning~\cite{siegwart2011amr}. Yet humans can often find their way to their destinations guided only by abstract, inaccurate representations of the environment - \eg{}, hand-drawn sketches or language-based directions - and approximate, semantic notions of their position. A key enabler of this skill is our ability to represent and navigate environments using \textit{navigational behaviours}, which are semantic action abstractions like \texttt{turn left} or \texttt{follow corridor}. Humans can use geometrically inaccurate maps or representations because these still capture paths in the environment abstractly, as sequences of navigational behaviours: \eg{} floor-plans allow us to infer the abstract sequences of \texttt{turning} and \texttt{corridor following} actions to take to reach a given room, despite their lack of realism. We can also perceive \textit{navigational affordances}~\cite{bonner2017affordances}, \ie{} the potential in the local environment for executing a navigational behaviour, and use them as non-metric, visual cues of our location: \eg{} observing that a nearby junction only affords us the chance to \texttt{turn left} and \texttt{go forward} can hint at our location in a building. We hypothesise that using navigational behaviours to represent and traverse environments imbues robots with the ability to navigate with limited metric and spatial information.

To test this hypothesis, we design a navigational behaviour-based robot system centred on the \textit{\samlong{}} (\sam{}), a topological representation comprising key places (nodes) joined by navigational behaviours (edges), that supports non-metric planning and localisation. In particular, we propose a learnable \textit{map-reading} pipeline to extract \sam{}s from various kinds of readily available, pre-existing 2D maps of the environment - \eg{} hand-drawn sketches and floor-plans. While many systems struggle to use such maps due to their metric inaccuracies and abstractness, ours instead uses the underlying \sam{}s encoded in these maps, allowing us to exploit this wealth of pre-existing map information.


\begin{figure}[!t]
    \centering
    \includegraphics[width=\linewidth]{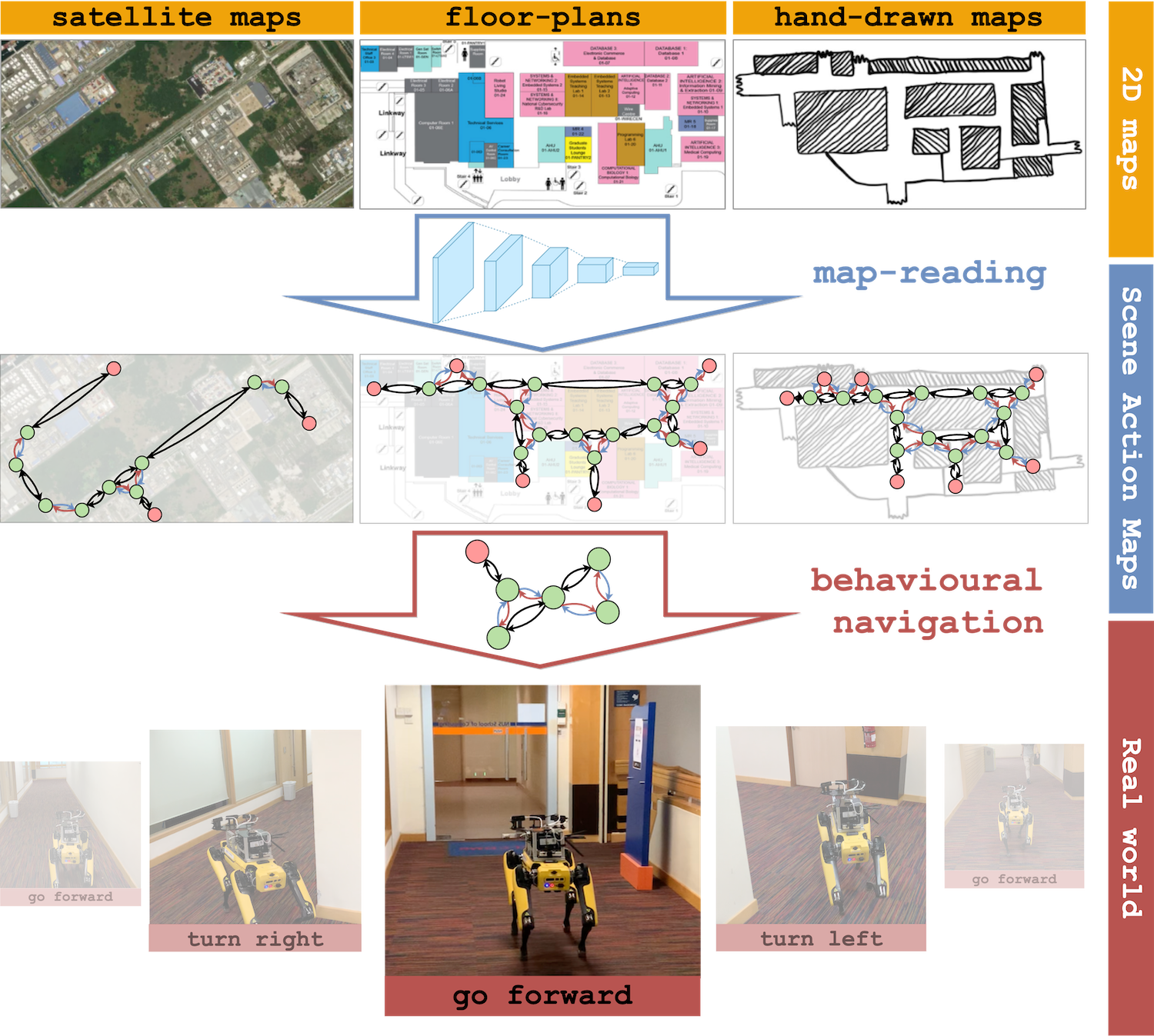}
    \caption{We propose a learnable \textit{map-reading} system that extracts \textit{Scene Action Maps} from pre-existing 2D maps, for behavioural navigation.}
    \label{fig:system_splash}
\end{figure}

Prior work in visual navigation has shown the practicality of learning human-like navigational behaviours~\cite{sepulveda2018behavioral, sorokin2022sidewalks, ai2022decision} and localising with their associated navigational affordances~\cite{chen2019graphnav}. Building on these, we implement a behavioural navigation stack employing \sam{}s and deploy it on a real robot to validate the usefulness of the extracted \sam{}s for navigation. In particular, we use the DECISION controller's obstacle-avoiding navigational behaviours~\cite{ai2022decision} and adapt Graph Localisation Networks~\cite{chen2019graphnav} for affordance-based localisation. We ``read'' \sam{}s from hand-drawn maps, floor-plans and satellite maps, and demonstrate that these extracted \sam{}s can be used for effective real-world navigation.

\section{Related Work}

Many robot navigation systems rely on accurate metric positioning and detailed geometric maps built with sensor data collected first-hand from the target environment~\cite{siegwart2011amr}. However, metric accuracy remains challenging to achieve in general: existing SLAM methods are sensitive to sensing conditions and robot dynamics, and can be difficult to extend to large-scale environments~\cite{cadena2016slam}. Further, by focusing on map-building with first-hand sensor data, these systems ignore the rich source of prior information for navigation provided by the wealth of pre-existing maps of manmade environments.

\subsection{Navigation without metric information}
To tackle the problems that metric inaccuracies pose for navigation, some works aim to improve the quality of metric mapping and localisation under challenging conditions~\cite{zhang2015vloam, ebadi2020lamp, chang2022lamp2}. In contrast, our work follows the approach of reducing the dependence of navigation on metric information.

A common strategy is to design navigation systems that only assume local metric consistency. Such systems typically navigate with topometric maps comprising independent metric submaps each with their own local reference frame, connected by a global graph~\cite{bosse2004slamatlas, modayil2004topometric, estrada2005hierarchicalslam, konolige2011topometricnav, schmid2021glocal}. Localisation and motion planning are done with respect to the submap the robot is currently in. While this helps to deal with issues like map corruptions owing to drift over long distances, it still depends on high metric accuracy locally within the submaps.

Purely topological approaches go further by avoiding metric information, instead representing the world relationally as a graph of perceptually significant places~\cite{lowry2015vpr}. A key challenge lies in grounding abstract topological routes to real-world metric paths and actions. Various action primitives, or behaviours, are used to bridge symbolic topological maps and the real world: Spatial Semantic Hierarchy and similar works use wall-following behaviours to travel between nodes~\cite{kuipers2000ssh, rawlinson2008topologynav, beeson2005voronoi}, while \cite{fraundorfer2007imagetopo} uses a visual servoing behaviour that takes image goals. However, these are often simple, handcrafted policies that can be brittle in the real world.

Deep learning provides a means of implementing robust behaviours, shown by its success in learning reactive visuomotor policies~\cite{sorokin2022sidewalks, zhu2017drl_nav, mirowski2017drl_nav, codevilla2018cil, gao2017inet, ai2022decision}. Recent works use this to imbue purely topological approaches with robust locomotion. SPTM~\cite{savinov2018sptm} and RECON~\cite{shah2021recon} build reachability graphs of the environment and learn image goal conditioned visuomotor policies to navigate between nodes. While these policies only consider the basic navigational affordance of reachability, \cite{sepulveda2018behavioral, chen2019graphnav} learn behaviour sets for indoor environments that exploit more semantically meaningful affordances: \eg{}, entering/exiting afforded by doors, path-following afforded by corridors etc. These affordances are associated with strong visual cues that help to guide behaviour execution, and are used in \cite{chen2019graphnav} to localise on a topological map.

Our \sam{}-based system is a topological, behavioural navigation system that builds on \cite{chen2019graphnav}. While \cite{sepulveda2018behavioral, chen2019graphnav} assume the graph is given, we propose to build \sam{}s from pre-existing maps. We also use the driving direction-based behaviours of \cite{gao2017inet, ai2022decision}, which apply to a wider range of environments. Lastly, while many prior works only evaluate their systems in simulation, we test ours in the real world.

\subsection{Navigation with pre-existing environment information}

While most SLAM approaches can only build maps for navigation from first-hand sensor data, some works build alternative navigation systems that leverage readily available, pre-existing maps for guidance, like hand-drawn sketches~\cite{boniardi2016sketch_nav, chen2020sketch_loc} or architectural floor-plans~\cite{boniardi2018cadfloorplan, boniardi2019floorplan_loc}. These are tailored to specific map types and make strong assumptions: the former assumes the sketch is a diffeomorphism of 2D LiDAR data, while the latter assumes mostly metrically accurate floor-plans. The I-Net system~\cite{gao2017inet} lowers accuracy needed in the latter by translating planned paths into sequences of path-following navigational behaviours, each locally guided by visual cues and affordances. However, it relies on metric localisation to switch between behaviours, and can fail under severe inaccuracies. Instead of tailoring navigation to specific map types, other works extract general intermediate representations for navigation: \eg{} road networks from aerial images/satellite maps~\cite{mattyus2017deeproadmapper, bastani2018roadtracer, li2019polymapper}.


This work proposes \sam{}s as a fully topological, general intermediate representation that can be extracted from a wide variety of readily available, pre-existing map types, ranging from sketches to floor-plans to satellite maps. Unlike prior approaches handcrafted for specific map types, our learnable ``map-reading'' system can be trained to extract \sam{}s from different map types for navigation.

\section{Methodology}

We consider the task of navigating to goals in environments the robot may not have seen or explored before. This naturally requires navigation with limited metric and spatial information, as lack of prior data means detailed geometric maps for planning and localisation may be unavailable. However, we assume access to readily available, pre-existing 2D maps of the environment like floor-plans, hand-drawn maps and satellite maps. Though they may be abstract and inaccurate, they retain information about the environment's navigational affordances useful for planning and localisation.

Some key challenges of this task include specifying goals, planning and localising with a range of abstract, inaccurate maps. Our approach is to extract a behavioural, topological graph of the environment from the maps, \ie{} a \textit{\samlong{}} (\sam{}), and navigate with it. We assume access to a set of navigational behaviours like DECISION~\cite{ai2022decision}, that are capable of local obstacle avoidance and diverse enough to allow us to reach most places in the target environment. Our \textit{offline map-reading} system is a learnable pipeline that, given a specific behaviour set, can be trained to extract \sam{}s from a variety of 2D maps. The \textit{online behavioural navigation} system takes in a goal specified on the \sam{}, plans a path over the \sam{} and executes it. Since we cannot depend on having accurate metric information, we use affordance-based localisation and learned navigational behaviours.


\subsection{\samlong{} }
\label{sec:map_design}

A \sam{} is a directed graph $\mathcal{G} = (\mathcal{V}, \mathcal{E})$. Each node $v\in\mathcal{V}$ stores a single label, marking it as a changepoint or destination node. Changepoint nodes denote locations at which the robot is \textit{afforded} the chance to transition between behaviours. For example, a changepoint may be placed just before a cross-junction, since the junction affords the robot the chance to switch from \texttt{going forward} to \texttt{turning left}. These nodes are extracted from pre-existing maps via offline map-reading. As the user may also want to set goals at places that are not changepoints, we allow them to specify destination nodes. The map-reading system takes in such nodes and includes them when forming the \sam{}.


Each directed edge $e\in\mathcal{E}$ stores a single label specifying the required navigational behaviour to move between its start and end nodes. The behaviours are drawn from a predefined set of $N$ navigational behaviours. In principle, \sam{}s can be adapted to use different behaviour sets.


We enforce the structural constraint that for any node $v$, each outgoing edge has a unique behaviour, implying $v$ has $\leq N$ outgoing edges. Intuitively this avoids the ambiguity of $v$ having multiple outgoing edges of the same behaviour type leading to different destinations, ensuring the problem of localising on a \sam{} is well-defined.

\begin{figure}
    \centering
    \includegraphics[width=\linewidth]{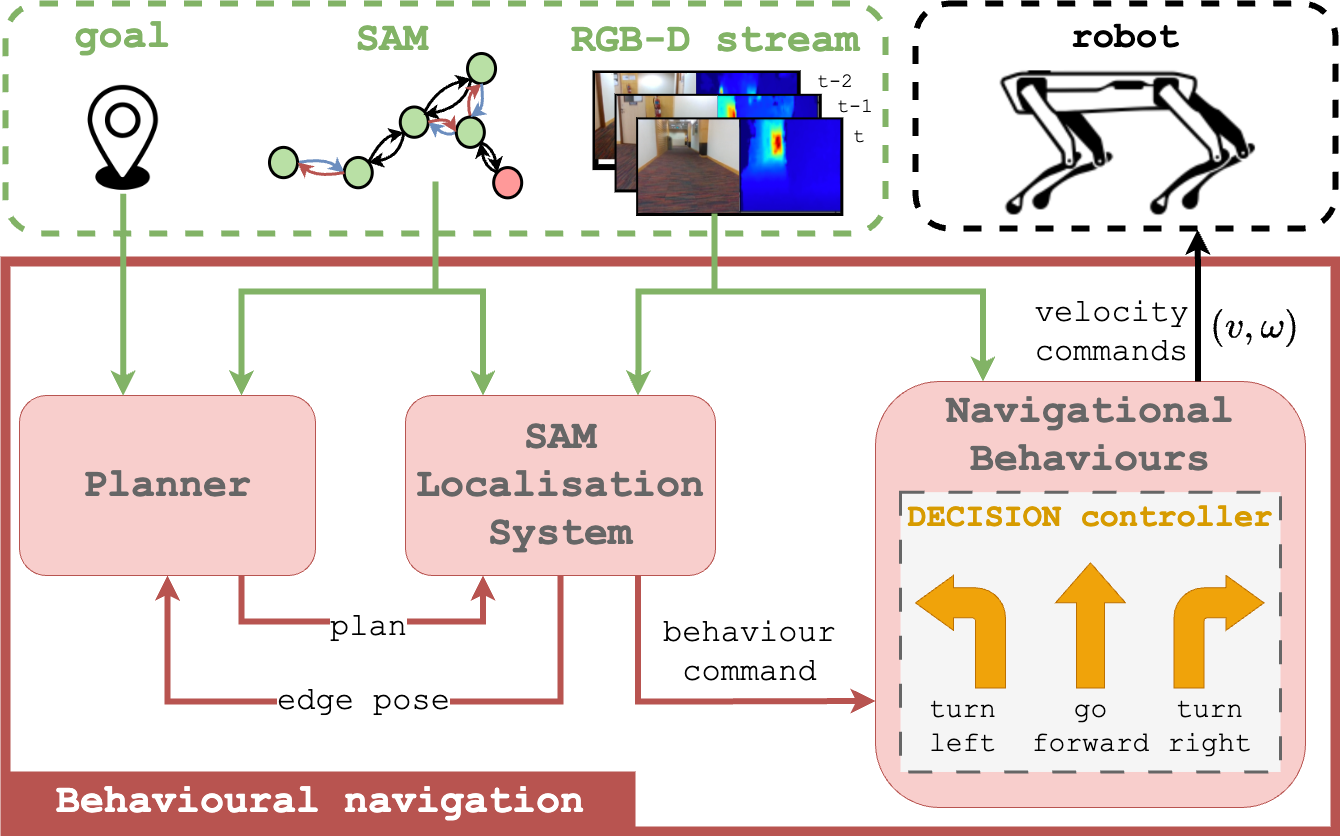}
    \caption{Overview of the online behavioural navigation system.}
    \label{fig:system_arch}
\end{figure}

\subsection{Online navigation with \sam{}s}
\label{sec:navigation}

To navigate to a specified goal, the \textit{online behavioural navigation} system (\autoref{fig:system_arch}) finds a valid path over the \sam{} with the planner, then executes it with learned navigational behaviours. The robot's position on the \sam{} is estimated by the \sam{} localisation system (SLS).

\textbf{Navigational behaviours:} These are a set of traversability-aware learned visual navigation policies that exploit navigational affordances. We use the DECISION controller~\cite{ai2022decision} which learns \{\texttt{turn-left}, \texttt{go-forward}, \texttt{turn-right}\} path-following behaviours with obstacle avoidance. These are meant to navigate along clearly defined paths, and may not handle areas like open spaces well. Regardless, our approach is not specific to this behaviour set, and can potentially be extended to other behaviours.


\textbf{\sam{} localisation system (\gls{}):} Like Graphnav~\cite{chen2019graphnav}, our system estimates the robot's pose as the \textit{edge} of the \sam{} it is currently on. This \textit{edge pose} is better defined than specifying the robot's pose as a node, since the robot may not always be near a node but is always executing an edge.


The \gls{} processes a sequence of depth images and a ``crop'' (local neighbourhood) of the \sam{} around the last edge pose. It estimates the current edge pose and triggers a behaviour switch command when nearing a changepoint. The \gls{} comprises a Graph Localisation Network (GLN) modified from Graphnav, a changepoint detector and edge pose predictor. The GLN encodes the image sequence into a feature vector capturing the local environment's structure and navigational affordances. It uses a GNN to match this vector with features encoding the \sam{}'s topology and behaviour information, assigning scores to edges based on their likelihood to be the edge pose. The edge pose predictor temporally smooths these ``edge scores'' and returns the best-scoring edge as the edge pose. Unlike Graphnav, we add a 3-layer MLP head to our GLN's last graph network block, that maps a given edge's features to a score indicating proximity to the changepoint at the edge's end. The changepoint detector temporally smooths and thresholds this ``changepoint score'' to determine changepoint proximity, and prompts a transition to the next planned behaviour if nearby.


We found explicit changepoint detection crucial for the robot to anticipate upcoming behaviour transitions. Slight lag in edge pose predictions from filtering, combined with our robot's higher speeds ($\sim$0.8m/s) compared to Graphnav's ($\sim$0.5m/s) meant our robot could overshoot the changepoint before a behaviour switch could occur. Predicting changepoint proximity provides a mechanism to anticipate this switch and change behaviours in time.

Let $\textbf{p}$ be a vector of unnormalized edge scores from the \sam{} crop, and $\textbf{q}_{gt}$ a vector of the ground truth edge pose's unnormalized changepoint scores. We train the GLN with cross-entropy losses on both $\textbf{p}$ and $\textbf{q}_{gt}$:

\begin{equation}
    \label{eqn:gln_loss}
    \mathcal{L}_{GLN}(\textbf{p}, \textbf{q}_{gt}) = \mathcal{L}_{xent}(\textbf{p}) + \lambda\mathcal{L}_{xent}(\textbf{q}_{gt}).
\end{equation}

\textbf{Planner:} Given a goal node, the planner searches over the \sam{} for a path from the current edge pose using Dijkstra. If a path exists, the planner returns a sequence of edges, and hence a sequence of behaviours to execute. During execution, the planner replans if the robot's edge pose deviates from the path. Since behaviours represent semantically meaningful, human-like actions, each plan is inherently interpretable.

We note that our changepoint detector lets us halt near goal nodes that are changepoints. For destination nodes, we heuristically halt after moving a preset distance along the edge leading to it. We do this since destinations are often near changepoints in indoor areas: \eg{}, reaching a room usually involves stopping shortly after entering and crossing the entrance changepoint. An alternative approach could have users label destination nodes with semantic location information (\eg{}, ``near desk in office'') to guide the robot on when/where to halt. 

\subsection{Offline map-reading} 
The diverse appearance of common, pre-existing 2D maps makes it challenging to handcraft general algorithms to extract \sam{}s from all map types. Instead, we propose a general pipeline that can be trained to extract \sam{}s from specific map types, given data of that map type. \sam{} prediction can be sequentially decomposed into \textit{node prediction}, followed by \textit{edge prediction} to connect the nodes with appropriate behaviours. We learn neural networks \cpnet{} and \emnet{} to aid in node and edge prediction. For a given map type and set of behaviours, if we can curate a dataset of such maps annotated with \sam{}s, we can train instances of \cpnet{} and \emnet{} to predict \sam{}s for that map type.


\textbf{Node prediction}: This takes in a 2D map $\mathcal{M}$ and yields $\mathcal{V}$, the set of nodes in the \sam{}. We learn \cpnet{}, a CNN that takes a patch from the map centred on a point, and outputs the likelihood that the point is a changepoint. \autoref{alg:predict_nodes} details how \cpnet{} is used to score points sampled from the map and cluster them to extract the set of changepoints in the map, $\mathcal{P}_c$. We return the set of nodes in the \sam{} $\mathcal{V} = \mathcal{P}_c\cup\mathcal{P}_d$, where $\mathcal{P}_d$ is the set of user-specified destination nodes. \cpnet{} is trained in two stages: we first learn a latent representation of changepoints using supervised contrastive learning (SupCon)~\cite{khosla2020supcon}, then add an MLP head and finetune the network to classify changepoints with a binary cross-entropy loss. Training data consists of positive samples drawn from a small neighbourhood around annotated changepoints, and negative samples from the rest of the map. We use SupCon as it allows us to exploit our data's positive/negative labels and learn a class-based representation. We find such contrastive representation learning to be essential for good convergence and achieving good changepoint predictions. 

 \begin{algorithm}[tb]
 \caption{Changepoint node prediction}\label{alg:predict_nodes}
 \begin{algorithmic}[1]
 \renewcommand{\algorithmicrequire}{\textbf{Input:}}
 \renewcommand{\algorithmicensure}{\textbf{Output:}}
 \REQUIRE 2D map $\mathcal{M}$, grid res $r$, changepoint threshold $\beta_{cp}$
 \ENSURE  Set of changepoints: $\mathcal{P}_c$
  \STATE $\mathcal{P} \leftarrow$ Grid-sampled points from $\mathcal{M}$ with resolution $r$
  \STATE $\mathcal{P}_{filtered} \leftarrow \{\}$;
  \FOR {each $p\in\mathcal{P}$}
  \STATE $m_p \leftarrow$ Crop image patch centred on $p$ from $\mathcal{M}$;
  \STATE $s_p \leftarrow$ \cpnet{}($m_p$), \ie{} likelihood $p$ is a changepoint;
  \IF {$s_p > \beta_{cp}$}
  \STATE $\mathcal{P}_{filtered} \leftarrow \mathcal{P}_{filtered}\cup\{p\}$;
  \ENDIF
  \ENDFOR
  \STATE Extract set of approximately convex clusters $\mathcal{C}$ from $\mathcal{P}_{filtered}$ with \textsc{GrowCompactClusters} from \cite{blochliger2018topomap};
 \STATE $\mathcal{P}_c \leftarrow$ Set of points with highest $s_p$ from each $c\in\mathcal{C}$;
 \RETURN $\mathcal{P}_c$ 
 \end{algorithmic} 
 \end{algorithm}

\textbf{Edge prediction}: This takes in a 2D map $\mathcal{M}$ and the set of nodes $\mathcal{V}$, and returns $\mathcal{E}$, the set of edges in the \sam{}. We compute $\mathcal{E} = \bigcup\limits_{v_i\in\mathcal{V}}E_{v_i} = \bigcup\limits_{v_i\in\mathcal{V}}\epfunc{}(v_i, \mathcal{M})$, where $\epfunc{}$ predicts the set of outgoing edges from $v_i\in\mathcal{V}$ while satisfying the \sam{} structural constraints. Intuitively, $\epfunc{}$ finds a partial assignment between behaviours and possible outgoing edges from $v_i$ under these constraints. Since $\epfunc{}$ must be learnable to adapt to different map types, we solve this optimisation differentiably with the Sinkhorn algorithm~\cite{cuturi2013sinkhorn}. This lets us predict the input score matrix to the Sinkhorn layer with a neural network, and learn to use visual information from $\mathcal{M}$ for edge prediction. For $N$ behaviours and $K$ nearest neighbour nodes to $v_i$, the score matrix has size $K\times N$. Intuitively the $(k,n)$-th element scores how likely it is that the \sam{} contains edge $e_{ik}$ with behaviour $n$. Like \cite{sarlin2020superglue}, we pad our score matrix with a ``dustbin'' row and column to allow for partial assignments. The Sinkhorn layer yields a soft assignment matrix $X$, that is thresholded during inference to get $E_{v_i}$. 

\begin{figure}
    \centering
    \includegraphics[width=0.9\linewidth]{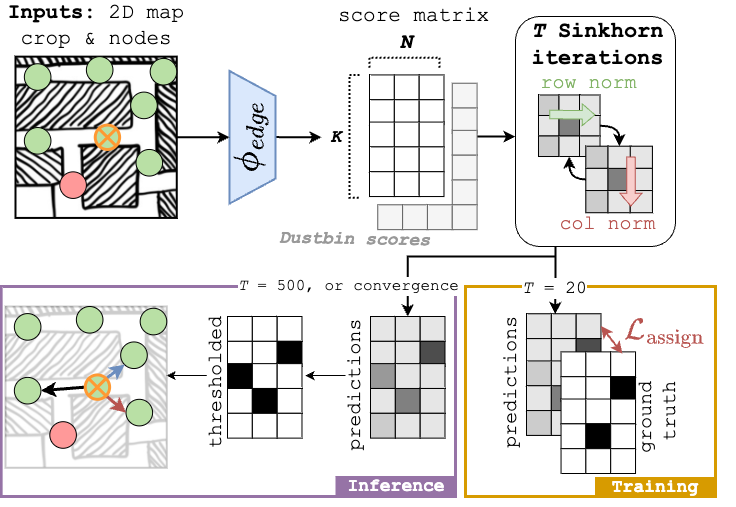}
    \caption{Applying $\epfunc{}$ to orange marked node: 1) predicts soft assignment matrix with \emnet{} and Sinkhorn, 2) thresholds to yield the outgoing edges.}
    \label{fig:emnet}
\end{figure}

Concretely, $\epfunc{}$ comprises \emnet{} (\autoref{fig:emnet}) followed by a Sinkhorn layer. \emnet{} is af CNN that predicts the score matrix from a $(3+K)$-channel input, where the first 3 channels are a crop of $\mathcal{M}$ around $v_i$, and the $(3+k)$th channel is a binary image with the $k$th neighbouring node's location rasterised on it. In practice \emnet{}'s fixed-size receptive field cannot easily handle environments where nodes can be very far apart, \eg{} at the ends of a long corridor. We tackle this with a multi-scale approach that learns separate \emnet{} networks to run $\epfunc{}$ at several scales. The edge predictions across scales are heuristically merged by preferring shorter pixel space edges. We learn a representation for \emnet{} with SupCon before finetuning it to predict edges. Given ground truth assignment matrix $A$, the loss for finetuning is:

\begin{equation}
\label{eqn:assign_loss}
\mathcal{L}_{assign} = -\frac{1}{\lVert A\rVert^2_F}\sum\limits_{n=1}^{N}\sum\limits_{k=1}^{K}a_{nk}\log(x_{nk}).
\end{equation}


\subsection{Implementation details}
\label{sec:implementation}
\cpnet{} is a MobileNetv2 with 4-layer MLP, trained with batch size 64, learning rates of $10^{-2}$ for SupCon and $10^{-4}$ for finetuning. \emnet{} is a MobileNetv2 with 3-layer MLP whose output is reshaped into the score matrix. We use learning rates/batch sizes of $10^{-2}/64$ for SupCon and $10^{-3}/32$ for finetuning. We train GLNs following Graphnav's method~\cite{chen2019graphnav}, and add data augmentation during training by randomly changing the assigned behaviour of a single edge in the \sam{} crop.

\section{Experiments}
Our experiments aim to answer the questions:
\begin{enumerate}
    \item How well can our \textit{offline map-reading} approach extract \sam{}s from varied 2D map inputs?
    \item Is our \textit{online behavioural navigation} system practical for real-world navigation? Do our proposed modifications to the GLN improve real-world performance?
    \item How well can the robot navigate using \sam{}s extracted by the \textit{offline map-reading} system?
\end{enumerate}
\subsection{Map-reading}

\begin{table}[tbp]
\caption{\textbf{Pr}ecision and \textbf{Re}call over various 2D map types for \textit{(A)} node prediction, \textit{(B)} edge prediction alone (ignoring behaviour correctness), \textit{(C)} edge and behaviour prediction}
\begin{center}
\begin{tabular}{@{\extracolsep{4pt}}ccccccc@{}}
\toprule
\textbf{}&\multicolumn{2}{c}{\textit{\hm{}}}&\multicolumn{2}{c}{\textit{\fp{}}}&\multicolumn{2}{c}{\textit{\sm{}}} \\
\cmidrule{2-3}\cmidrule{4-5}\cmidrule{6-7}
\textbf{Tasks} & \textbf{Pr} & \textbf{Re} & \textbf{Pr} & \textbf{Re} & \textbf{Pr} & \textbf{Re} \\
\midrule
\textit{(A)} &0.848 &0.975 &0.732 &0.779 &0.865 &0.621   \\
\textit{(B)} &0.754 &0.605 &0.820 &0.643 &0.863 &0.751 \\
\textit{(C)} &0.667 &0.535 &0.630 &0.494 &0.761 &0.662 \\
\bottomrule
\end{tabular}
\label{tab:map_reading_perf}
\end{center}
\end{table}

We collect data for the 3 map types in \autoref{fig:system_splash}: hand-drawn maps (\hm{}) and floor-plans (\fp{}) of campus buildings, and satellite maps (\sm{}) of industrial areas. \sam{}s are manually annotated for maps in \hm{} and \fp{} datasets. \sam{}s are annotated using OpenStreetMap road/junction information for \sm{} maps. We train a separate instance of our map-reading module for each map type. To answer \textbf{Q1}, we test on held-out datasets: \hm{}/\fp{} each have 4 maps with each map having a mean of 27 nodes and 64 edges, and \sm{} has 1 large map with 137 nodes and 414 edges. We compute precision and recall for 3 tasks: (A) predicting nodes/changepoints, (B) predicting edges alone (disregarding behaviour correctness), and (C) predicting edges along with their associated behaviours. Intuitively (B) shows how well the environment's structure and connectivity is captured. (C) further checks each edge's assigned behaviour against human-annotated maps. Results are presented in \autoref{tab:map_reading_perf}.

Our node prediction performs well at predicting changepoints across all map types. Qualitatively, \cpnet{} is able to reliably capture visual features in the maps, like junctions or turnings, that can indicate a changepoint when using the DECISION behaviour set. Failures mainly occur in open areas where the environment's structure is less well-defined, leading to more false positives and negatives. The comparatively lower recall score for \sm{} is mostly due to features like junctions being occluded by tall buildings in densely built-up areas, inducing more false negatives.


Our edge prediction performs well on task (B), particularly on \sm{} due to the rich visual information inherent in satellite maps. The lower recall scores indicate that $\epfunc{}$'s main limitation is its occasional failure to identify valid edges. Lower performance on task (C) compared to (B) suggests that while \emnet{} can learn reachability between nodes well, it is significantly more challenging to learn the right visual features needed to assign correct behaviours. This is supported by the observation that most failures involve a \texttt{go-forward} behaviour being wrongly assigned as a turning behaviour and vice versa.

We connect node and edge prediction, and generate \sam{}s end-to-end in \autoref{fig:example_maps}. Our method can trace out connected graphs that capture the topology of the map reasonably accurately. While there is some noise in the predicted \sam{}s - in the form of missing changepoints, edges with mislabelled behaviours etc. - we demonstrate that these \sam{}s can still be effectively used for behavioural navigation.

\subsection{Indoor robot  navigation}
\begin{figure}
    \centering
    \includegraphics[width=\linewidth]{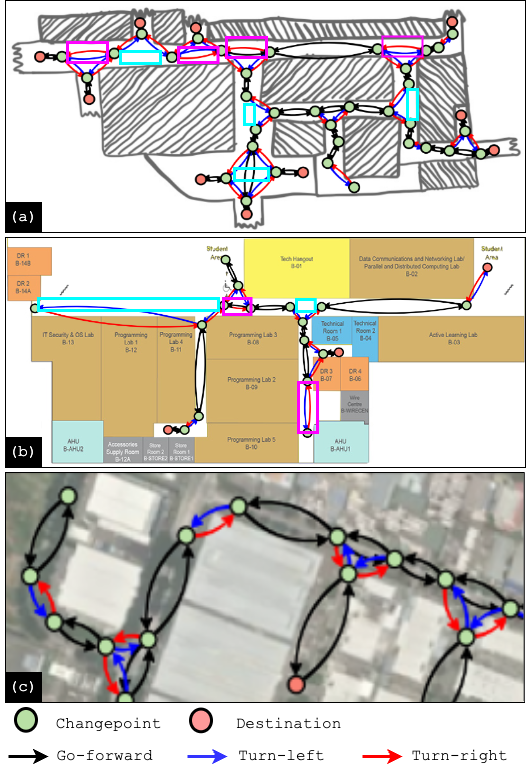}
    \caption{\sam{}s extracted from (a) hand-drawn maps, (b) floor-plans, (c) satellite maps. The \sam{}s mostly capture the behaviours and environment structure accurately, apart from occasional errors (circled on maps) like \textcolor{magenta}{confusing \texttt{go-forward} and turning edges}, or \textcolor{cyan}{missing edges}.}
    \label{fig:example_maps}
\end{figure}

\begin{figure}
    \centering
    \includegraphics[width=\linewidth]{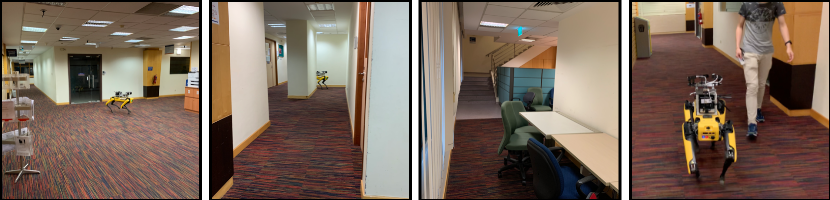}
    \caption{Navigating our test environment is a challenge due to (left-to-right) wide open spaces, complex junctions, cluttered areas and dynamic obstacles.}
    \label{fig:test_env}
\end{figure}

We run the \textit{online behavioural navigation} system on a Boston Dynamics Spot robot with an AGX Xavier and 3 Realsense d435i RGB-D cameras with combined 140$^{\circ}$ FoV. 

Our navigation tests are split into 3 difficulty levels based on the number of changepoints the robot must transit through: \easy{} (2-3 changepoints over 10-25m), \med{} (4-5 changepoints over 30-50m), and \hard{} (6-10 changepoints over 50-100m). All changepoints in the \sam{} are included in the \easy{} and \med{} routes. \hard{} routes involve a circuit around the entire building level being tested. Graphnav~\cite{chen2019graphnav} defines difficulty differently: their corresponding levels contain paths with 1-10 nodes, 11-20 nodes and 21-30 nodes. Their paths are shorter despite having more nodes (median/maximum lengths of 30/60m). Since nodes are closely packed in their environment, their robot easily accumulates ample topological features to localise. In contrast, our environments contain long corridors with sparse junctions and changepoints, forcing our GLN to localise with sparser features. Our real-world tests also include complex environment structure and distractors (\autoref{fig:test_env}), making navigation harder than in Graphnav's simulation tests.

Like Graphnav, we assess navigation performance using Success Rate (\textit{SR}) and Plan Completion (\textit{PC}). \textit{PC} is the proportion of nodes on the path successfully transited through. Based on \textit{SR} and \textit{PC}, we define two finer-grained sets of metrics: \textit{\srnav{}} and \textit{\pcnav{}} to capture overall system performance, and \textit{\srloc{}} and \textit{\pcloc{}} to capture the performance of the high-level navigation system - \ie{} planner and localisation - in isolation. For the \textit{Nav} metrics, failure is if an incorrect behaviour is issued or a behaviour fails. For \textit{HL}, failure only occurs if an incorrect behaviour is issued: a controller failure is manually corrected by the operator.


To answer \textbf{Q2}, we find that the behavioural navigation system using our proposed \sam{} Localisation System (\glne{}) performs well on all routes when given high-quality, human-annotated (HA) \sam{}s, specifically hand-drawn maps (\hmgt{}) and floor-plans (\fpgt{}). Our \glne{} outperforms Graphnav's localisation system \glnh{}, which combines their vanilla GLN with heuristics that use the edge probability distribution to decide when to switch behaviours. This is due to \glnh{}'s failure to switch behaviours in time, which \glne{} mitigates with changepoint proximity detection.

We answer \textbf{Q3} by showing that effective behavioural navigation is possible with ``noisy'' predicted \sam{}s, which may contain defects such as edges labelled with the wrong behaviours or missing nodes/edges (see \autoref{fig:example_maps}). We evaluate both \glne{} and \glnea{} on noisy \sam{}s, with \glnea{} using a GLN trained with our proposed data augmentations for enhanced noise robustness. We draw 2 conclusions from \autoref{tab:nav_maps}. Firstly, navigation performance experiences minimal adverse effects when substituting noisy predicted \sam{}s for human-annotated \sam{}s, indicated by the fact that \glne{} systems see at most a small drop in \textit{PC} between human-annotated and predicted \sam{}s. Empirically, \glne{} and \glnea{} seem robust to the common modes of noise - \ie{} missing edges at a junction or confusing \texttt{go-forward} and turning behaviours - and is often able to use the remaining correct topological features to localise and navigate. Secondly, our data augmentation improves localisation and navigation on predicted \sam{}s containing noise and artifacts. On predicted \sam{}s, \glnea{} generally outperforms other test settings, and even surpasses \glne{} on human-annotated \sam{}s. Overall, \glnea{} shows promising performance even on \hard{} routes of up to 100m with multiple changepoint transitions, affirming the feasibility of predicting \sam{}s from 2D maps to localise and navigate in the real world.


\begin{table}[tbp]
\caption{Comparing navigation performance on human-annotated (HA) \sam{}s with \glnh{} (Graphnav) and our \glne{}, and on noisy, predicted \sam{}s with \glne{} and \glnea{}}
\begin{center}
\begin{tabular}{clcccc}
\toprule
\multicolumn{2}{c}{\textbf{Test settings}} & \textit{\srloc{}}& \textit{\pcloc{}}& \textit{\srnav{}}& \textit{\pcnav{}} \\
\midrule
\multirow{8}{*}{Easy}& \hmgt{} (\textit{\glnhtab{}}) &45.5 &56.8 &45.5 &52.3 \\
&\hmgt{} (\textit{\glnetab{}}) &70.0 &85.0 &70.0 &85.0 \\
&\hm{} (\textit{\glnetab{}}) &70.0 &85.0 &70.0 &85.0 \\
&\hm{} (\textit{\glneatab{}}) &\textbf{80.0} &\textbf{90.0} &\textbf{80.0} &\textbf{90.0} \\
\cmidrule{2-6}
& \fpgt{} (\textit{\glnhtab{}}) &12.5 &43.8 &12.5 &43.8 \\
& \fpgt{} (\textit{\glnetab{}}) &66.7 &77.8 &66.7 &77.8 \\
& \fp{} (\textit{\glnetab{}}) &\textbf{88.9} &\textbf{94.4} &\textbf{88.9} &\textbf{94.4} \\
& \fp{} (\textit{\glneatab{}}) &68.8 &78.1 &62.5 &71.9 \\
\midrule
\multirow{8}{*}{Med}& \hmgt{} (\textit{\glnhtab{}}) &12.5 &21.9 &12.5 &21.9 \\
&\hmgt{} (\textit{\glnetab{}}) &50.0 &\textbf{87.5} &50.0 &\textbf{81.3} \\
&\hm{} (\textit{\glnetab{}}) &\textbf{75.0} &84.4 &\textbf{62.5} &75.0 \\
&\hm{} (\textit{\glneatab{}}) &\textbf{75.0} &\textbf{87.5} &\textbf{62.5} &75.0 \\
\cmidrule{2-6}
& \fpgt{} (\textit{\glnhtab{}}) &0.0 & 18.8 &0.0 &12.5 \\
& \fpgt{} (\textit{\glnetab{}}) &\textbf{37.5} &62.5 &12.5 &43.8 \\
&\fp{} (\textit{\glnetab{}}) &22.2 &56.7 &22.2 &56.7 \\
&\fp{} (\textit{\glneatab{}}) &\textbf{37.5} &\textbf{65.6} &\textbf{37.5} &\textbf{65.6} \\
\midrule
\multirow{6}{*}{Hard}& \hmgt{} (\textit{\glnhtab{}}) &0 &12.5 &0 &12.5 \\ 
&\hmgt{} (\textit{\glnetab{}}) &\textbf{50.0} &75.0 &\textbf{50.0} &75.0 \\
&\hm{} (\textit{\glnetab{}}) &0 &71.7 &0 &71.7 \\
&\hm{} (\textit{\glneatab{}}) &\textbf{50.0} &\textbf{85.0} &\textbf{50.0} &\textbf{85.0} \\
\cmidrule{2-6}
& \fpgt{} (\textit{\glnhtab{}}) &0 & 15.8 &0 &15.8 \\
& \fpgt{} (\textit{\glnetab{}}) &0 &73.9 &0 &73.9 \\
&\fp{} (\textit{\glnetab{}}) &0 &33.3 &0 &33.3 \\
&\fp{} (\textit{\glneatab{}})&\textbf{50.0} &\textbf{80.0} &\textbf{50.0} &\textbf{80.0} \\
\bottomrule
\end{tabular}
\label{tab:nav_maps}
\end{center}
\end{table}

\section{Conclusion}
We introduced \textit{\samlong{}s}, a behavioural topological representation for navigation. We recognise that common, pre-existing maps like floor-plans often encode information on navigational affordances and behaviours, and propose a ``map-reading'' system to extract \sam{}s from such maps. We also show effective real-world navigation with \sam{}s extracted from sketches and floor-plans. 

\sam{}s make a trade-off: by being constrained to specific behaviour sets (and hence robot dynamics) they reduce reliance on metric information. In contrast, geometric maps need accurate data and cannot be built from abstract inputs, but represent the world richly enough to enable navigation with a wide range of robot dynamics. In future work, we intend to test our system in outdoor environments and incorporate richer sources of information into \sam{}s.

\section*{Acknowledgment}
This research is supported by Agency of Science, Technology \& Research (A*STAR), Singapore  under its National Robotics Program (No. M23NBK0053), and also the DSO National Laboratories' Graduate Fellowship.

\bibliographystyle{IEEEtran}
\bibliography{references}

\addtolength{\textheight}{-12cm}   







\end{document}